\documentclass{bmvc2k}

\title{STARS: Zero-shot Sim-to-Real Transfer for Segmentation of Shipwrecks in Sonar Imagery}

\addauthor{Advaith Venkatramanan Sethuraman}{advaiths@umich.edu}{1}
\addauthor{Katherine A. Skinner}{kskin@umich.edu}{1}

\addinstitution{
 Department of Robotics\\
 University of Michigan\\
 Ann Arbor, Michigan, USA
}

\runninghead{Sethuraman and Skinner}{STARS}
\usepackage{graphicx}
\usepackage{pifont}
\usepackage{wrapfig}
\usepackage{amsmath}
\usepackage{multicol}
\usepackage{amsfonts}
\usepackage{enumitem}
\usepackage{svg}

\newcommand{\cmark}{\ding{51}}%
\newcommand{\xmark}{\ding{55}}%

\begin{document}

\maketitle
%intro structure 
%1. auvs get this data 
%2. No large datasets, but we can generate synthetic data 
%3. sadly there is a big sim2real gap that motivates research in zero-shot sim2real transfer for this field 
%4. Our method STARS addresses this in a unique way, improving performance 
\begin{abstract}
%Large area search and survey missions conducted by autonomous underwater vehicles (AUVs) often use side scan sonar as a sensing modality. However, identifying targets of interest in large amounts of sonar imagery is a time consuming task that requires domain expertise. 
%Application of traditional supervised deep learning techniques to automate this task remains difficult due to the lack of large labeled datasets for side scan sonar imagery and the limited frequency of target objects observed in real datasets. %Although recent advances in sonar simulation can produce large synthetic datasets for training, a significant sim-to-real gap exists between synthetic sonar imagery and real sonar imagery collected from real deployments. Since collecting even small sonar datasets for unsupervised domain adaptation techniques is expensive and time consuming, this problem motivates advances in vision algorithms that can better transfer from simulation directly to reality in a \textit{zero-shot} manner. 
In this paper, we address the problem of sim-to-real transfer for object segmentation when there is no access to real examples of an object of interest during training, i.e. \textit{zero-shot sim-to-real transfer for segmentation}. We focus on the application of shipwreck segmentation in side scan sonar imagery. Our novel segmentation network, STARS, addresses this challenge by fusing a predicted deformation field and anomaly volume, allowing it to generalize better to real sonar images and achieve more effective \textit{zero-shot sim-to-real transfer} for image segmentation. We evaluate the sim-to-real transfer capabilities of our method on a real, expert-labeled side scan sonar dataset of shipwrecks collected from field work surveys with an autonomous underwater vehicle (AUV). STARS is trained entirely in simulation and performs zero-shot shipwreck segmentation with no additional fine-tuning on real data. Our method provides a significant \textbf{20\%} increase in segmentation performance for the targeted shipwreck class compared to the best baseline. 
% % This paper presents a novel deep learning framework to automate the segmentation of sites of shipwreck sites from sonar imagery collected by an autonomous underwater vehicle (AUV). 
% A major challenge faced by traditional computer vision algorithms for this task is the lack of large labeled datasets for supervised network training. To overcome the lack of large labeled datasets, we propose a  method for generating diverse synthetic side scan sonar data. Still, a sim-to-real gap exists between synthetic sonar imagery and real sonar imagery collected from real deployments. Our novel segmentation network, STARS, addresses this gap by fusing a predicted deformation field and anomaly volume, allowing it to generalize better to real sonar images to achieve zero-shot sim-to-real-transfer for segmentation. We evaluate the segmentation performance of our method on an expert-labeled real side scan sonar dataset of shipwrecks collected from field work surveys with an AUV.
% %in Thunder Bay National Marine Sanctuary in Lake Huron. 
% Our method provides a significant \textbf{20\%} increase in segmentation IOU performance for the targeted shipwreck class compared to the best baseline. STARS is trained entirely in simulation and performs zero-shot shipwreck segmentation with no additional fine-tuning on real data.
\end{abstract}

%-------------------------------------------------------------------------
\section{Introduction}
%1. motivation for shipwrecks 
%2. go-to solution for solving segmentation problems, present lack of data in this field 
%3. go-to solution for lack of labeled data: simulation, present sim2real gap problem and make it specific to this problem
%4. go-to solutions for closing the sim-to-real gap and why this problem is STILL challenging. this is the "research gap"
%5. our proposed solution for addressing these research gaps
%Preservation and protection of significant historical shipwrecks have been a concern for national policymakers \cite{pres}.% Ninety-three shipwrecks have been discovered in the Thunder Bay National Marine Sanctuary (TBNMS) in Lake Huron, Michigan. However, it is estimated one hundred sites are yet to be uncovered. 
%Underwater search and discovery plays a vital role in search and rescue operations, historical preservation efforts, and national defense initiatives. %However, locating submerged objects, such as shipwrecks, downed airplanes, and mines, is a challenging task. 
%The latest techniques for underwater search leverage autonomous underwater vehicles (AUVs) equipped with side scan sonar. 
Autonomous underwater vehicles (AUVs) equipped with sonar systems have demonstrated potential to carry out efficient, cost-effective large area surveys of marine environments to locate submerged objects, such as shipwrecks, downed airplanes, and underwater mines. Still, identification of submerged objects from sonar data is currently a manual process relying on expert knowledge for interpretation, which can take many months to complete. %, often requiring multiple surveys to verify potential new discoveries. % search for targets of interest such as fallen aircraft, underwater mines, or marine archaeological sites.  %The conventional approach to finding these sites is to perform surveys using ship-mounted acoustic sensors over large search areas. Then, experts must go through the vast amount of imagery collected and identify areas of interest. Finally, human divers are sent to confirm the wrecks. This process is time-consuming and resource intensive.

On land, deep learning algorithms are able to leverage large, publicly available training datasets to achieve impressive performance on automated computer vision tasks such as object detection and segmentation from optical imagery~\cite{ade20k, cityscapes}. 
% Leveraging large, publicly available labeled datasets for training, deep learning has demonstrated impressive performance on automated computer vision tasks such as object detection and segmentation from optical imagery~\cite{ade20k, cityscapes}.
% This can reduce labor in manually identifying targets from images. %In this paper, we focus on semantic segmentation in side scan sonar imagery (SSS), which is widely used on Autonomous Underwater Vehicles (AUVs) for high altitude, large area surveys. %SSS provides high resolution imagery with high speckle noise, due to the scattering of sound underwater. 
%need to motivate the data restriction here and present possible solutions to finding shipwrecks still. 
%1. simulation is an option
% - talk about domain adaptation, but with a class restriction since no real shipwrecks
%2. unsupervised anomaly detection
%shipwrecks in real life can look very different from simulation
%Deep neural networks have provided significant leaps in performance for semantic segmentation algorithms in a variety of real world environments \cite{ade20k, cityscapes}. 
% However, state-of-the-art methods rely on access to large, labeled datasets for training. 
Unfortunately, domain-specific barriers like security concerns, cost of collection, and difficulty labeling sonar data prevent the creation of large, publicly available training datasets for side scan sonar.
%65\% of the Earth's surface remains unmapped. This is due to the fact that 70\% of our planet is covered by water, a medium that cannot be easily imaged by long-range, large-area remote sensing methods like satellite imagery. Rather, imagery of marine terrain must be captured in short-range, small-area surveys using acoustic sensors like sonar. This poses a major scale problem for image understanding underwater: 
%Unfortunately, there is a lack of large, publicly available labeled datasets for sonar imagery. 
Furthermore, even when data is available, real examples of specific target objects may not be present due to limited frequency of appearance. This motivates the need to develop new machine learning methods capable of performing accurate object detection and segmentation in spite of significant data restrictions for network training.

% State-of-the-art deep learning methods still rely on large amounts of labeled training data. For many applications across marine robotics, it can be extremely expensive to collect and label these training datasets.Specifically, large, labeled public sonar datasets are not available because of the cost of collection and security concerns. %To remove the reliance on real, labeled sonar data and accelerate the application of deep learning to marine robotics, we investigate segmentation methods that can learn completely in simulation. 

\begin{figure}[t]
    \centering
    \includegraphics[width=\textwidth]{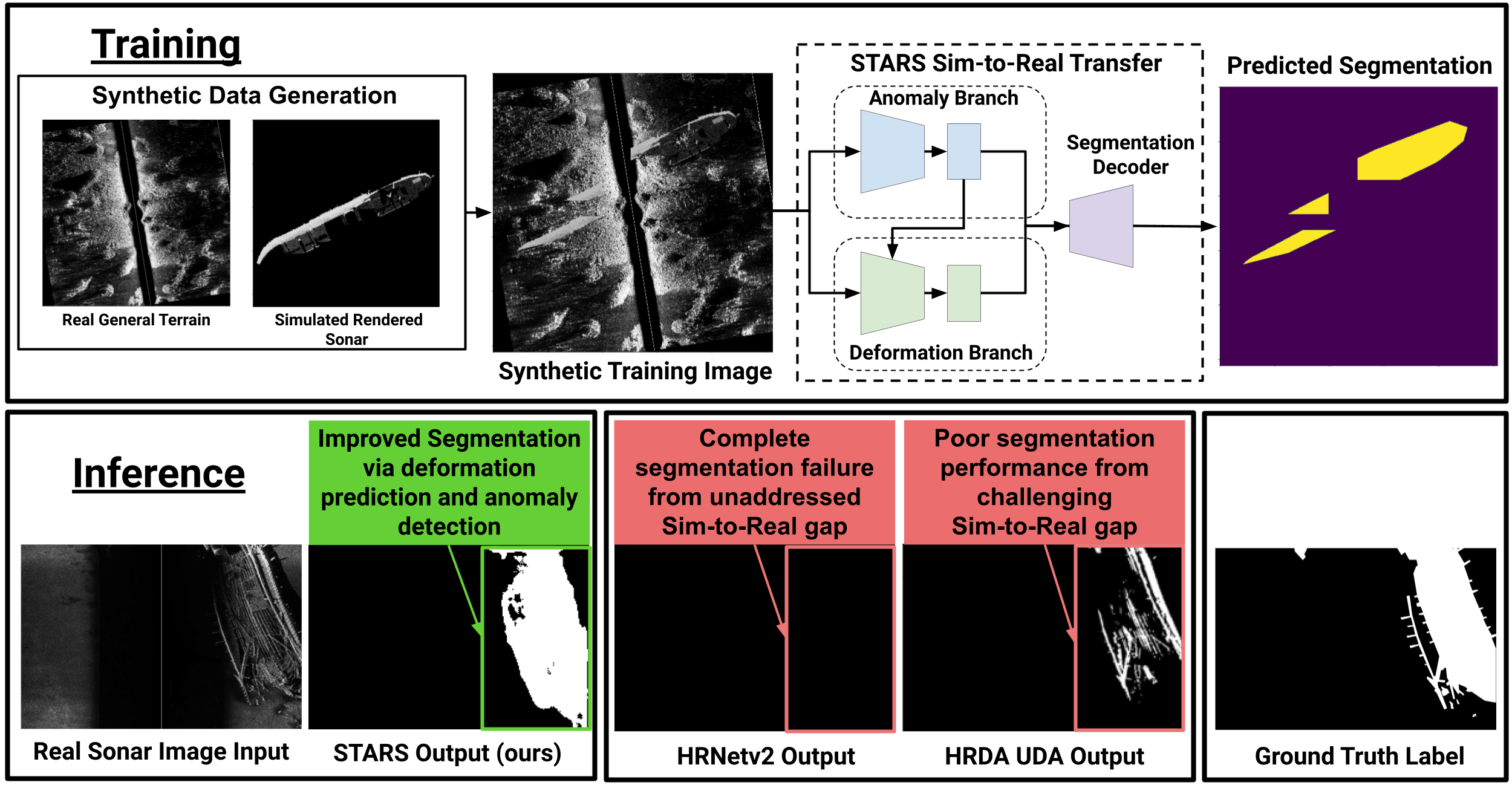}
    \caption{The primary problem this work addresses is the sim-to-real gap between synthetic training data and real testing data for side scan sonar images. We find that naively training segmentation (HRNetv2) and domain adaptation models (HRDA) on synthetic data does not transfer effectively to diverse and unstructured real sonar data. Our method STARS addresses this research gap with the novel fusion of a predicted deformation field and anomaly volume. STARS takes a step towards more accurate segmentation in never-before-seen environments. \label{intro}}
    \vspace{-10mm}
\end{figure}

% The primary challenges for semantic segmentation in SSS imagery are: 1) there is a limited amount of training data available for marine applications and 2) there is high variance of object appearance dependent on site condition, sensor configuration, and environmental characteristics, leading to bad generalization. Current approaches for object identification in sonar imagery rely heavily on real, labeled datasets for fine-tuning or data augmentation \cite{ayoung1, geosciences9040159, volov5}. However, this approach may not always be viable given the cost of collection and labeling. Furthermore, due to the rare nature of shipwrecks, we cannot assume access to a large dataset of real shipwreck images at train time. 

Simulation is useful for generating data that is difficult and expensive to collect and it can be leveraged to generate additional examples of rare objects or events. %Techniques like domain adaptation have been developed to improve the performance of networks trained in simulation and tested on real data. 
%Simulation has been used to provide large amounts of labeled data for training deep learning algorithms and can be leveraged to generate additional examples of rare objects or events. 
However, as shown in Fig. \ref{intro}, naively training state-of-the-art supervised segmentation methods on simulated data fails due to the \textit{sim-to-real} gap between the synthetic data and real data, which hinders model performance at test time \cite{rss_sethuraman, adda}. %Complex environmental factors like water temperature, currents, particulate concentration, and terrain properties greatly affect the sonar image formation process and are non-trivial to model. %Shipwreck segmentation poses another unique challenge: each site has varying levels of destruction and environmental degradation. Sites like the one shown in Fig. \ref{intro} can pose a great challenge to perception algorithms. 
%Moreover, representing the full diversity of possible sites in synthetic data can be an intractable task. %, motivating a network design that can minimize this \textit{sim-to-real} gap.
% Secondly, perception systems that operate in unstructured environments can encounter scenes that are completely different from those seen during training. This \textit{distribution shift} between training data and deployment can be catastrophic and lead to failure in perception systems \cite{dist_shift}.  In an exploratory setting, there is no guarantee that a training set of shipwrecks will capture the true diversity of novel sites seen during deployment.  
%should briefly talk about how sim2real gap has been addressed before and why its not applicable to this task. 
%Current approaches to closing the \textit{sim-to-real} gap include domain randomization, domain adaptation, and learning domain invariant features \cite{ adda,sim2realarticle, domrand, guda}. Domain randomization aims to make models more generalizable by randomly changing parameters like color, texture, lighting, and object poses \cite{domrand}. However, modelling the complex deformations of a shipwreck through domain randomization is a difficult task \cite{sim2realarticle}. 
Techniques like domain adaptation have been developed to improve the performance of networks trained in simulation and tested on real data. 
%Domain adaptation takes data from a source domain (synthetic data) and adapts a model to perform inference in a target domain (real data) \cite{adda}. 
Still, many domain adaptation methods require examples of objects of interest (i.e. shipwrecks) in the target domain. %Even with access to real examples, network generalization can be limited by intra-class diversity. This motivates the need to develop new methods for sim-to-real transfer for scenarios where real examples of target objects are not available at training time. %capable of performing accurate object detection and segmentation in spite of significant data restrictions for network training.%Finally, networks can learn the salient features that don't change between domains, leading to better performance in the target domain \cite{guda}. However, learning invariant features also assumes access to real shipwrecks. 
%The lack of labeled training data motivates unsupervised approaches. 
% Anomaly detection is a popular solution when access to labeled training data is restricted and the configuration of objects we are interested in is ill-defined \cite{patchcore, dest}. These algorithms can use normal examples for training and detect anomalies during test time. However, anomaly detection algorithms can have high false-positive rates when used in unstructured environments, such as those underwater. Additionally, these methods detect general anomalies and do not provide insight on the presence of specified target objects.

In this paper, we address the problem of sim-to-real transfer for object segmentation when there is no access to real examples of an object of interest during training, i.e. \textit{zero-shot sim-to-real transfer for segmentation}. 
%We focus on an application relevant to marine robotics and marine archaeology -- detection and segmentation of novel shipwreck sites in side scan sonar imagery.
Our main contributions are: 
\begin{itemize}[noitemsep,nolistsep]
   
    \item We introduce a novel zero-shot sim-to-real segmentation framework,  \textbf{STARS}: \textit{Sonar Target Anomaly Recognition for Segmentation} \footnote{\url{https://umfieldrobotics.github.io/STARS.github.io/}}, that exploits cues from anomaly detection and deformation prediction for improved segmentation performance. 
    \item We propose a simple but effective synthetic side scan sonar image generation method that produces diverse and randomized debris fields to support training for sim-to-real transfer for side scan sonar imagery.
    \item We present extensive qualitative and quantitative evaluation on a real side scan sonar dataset consisting of 220 scans of 14 distinct shipwreck sites collected with an AUV in Thunder Bay National Marine Sanctuary (TBNMS), which includes detailed and high-resolution labels generated by an expert marine archaeologist. 
\end{itemize} %We validate STARS with extensive experiments using our dataset of real sonar images of shipwrecks collected with an AUV in Thunder Bay National Marine Sanctuary (TBNMS).
Through experiments, we show that our STARS provides a significant \textbf{20\%} improvement in segmentation performance for the shipwreck class over state-of-the-art baselines. %, including methods addressing unsupervised anomaly detection, semantic segmentation, unsupervised domain adaptation, and salient object detection.

\section{Related Work}
% \subsection{Object Detection in Sonar Imagery}
% Prior work has focused on supervised training on real sonar data, including finetuning on small datasets~. 
\subsection{Object Detection and Segmentation in Sonar Imagery} Recent work on object detection and segmentation from sonar imagery has leveraged machine learning. The majority of work in object detection for sonar imagery involves fine-tuning existing object detection algorithms on real, labeled sonar imagery~\cite{8604879,geosciences9040159, SSSEG, Nayak2019MachineLT}. %It is important to note that these works have access to expert-labeled side scan sonar data for training, which is expensive and time consuming to collect. 
The unique nature of side scan sonar imagery has also motivated the development of specialized network architectures \cite{jmse8080557, baseline2}, but these networks still require access to labeled datasets. %Burguera and Bonin-Font use an encoder/decoder architecture to perform segmentation of side scan images \cite{jmse8080557}, whereas Yang et. al use a multi-channel segmentation network to segment terrain classes in side scan sonar \cite{baseline2}. Both networks are trained using supervised learning on labeled datasets. 
% Recently, Valdenegro-Toro \textit{et al.} \cite{objectnessdet} proposed to adopt a sliding window approach based on the window patch's objectness score to detect marine debris from forward-looking sonar images. Although effective approaches, these methods require manually labeling small datasets. 
Due to data restrictions for target objects, our work instead aims to perform segmentation of shipwreck sites without access to real labeled examples during training. 

% \subsection{Synthetic Sonar Data Generation}
%\subsection{Synthetic Sonar Data Generation} 
Recent work has focused on leveraging simulation to generate realistic side scan sonar imagery to overcome data limitations. The image formation process for side scan sonars can be approximated using ray-tracing and does not require access to real, labeled sonar data \cite{sssrend}. However, the lack of diverse terrains and consideration of environmental factors leads to a \textit{sim-to-real} gap between the synthetic sonar data and real sonar data that can hinder model performance on real data at test time \cite{rss_sethuraman, adda, simsss}. Current approaches to improving realism in synthetic sonar data generation include Generative Adversarial Networks~\cite{ayoung3,9137342}, style transfer~\cite{ayoung2,ayoung1}, ray-tracing~\cite{sssrend}, and image composition techniques \cite{mines, sss_compos, Misra-2017-103478}. %For example, Kim et al. use a Pix2Pix image translation network trained on real, labeled data to convert from binary masks to objects in sonar imagery \cite{ayoung3}. \cite{ayoung2} use depth images generated in simulation then use a style transfer network to emulate different environments like pool and ocean. 
%Another method for sonar image synthesis is ``cutting and pasting" \cite{mines,sss_compos, Misra-2017-103478}. 
For example, \cite{sss_compos} uses style transfer networks and cutting and pasting to produce synthetic side scan sonar images. Although effective, this method also requires real examples of objects in side scan sonar imagery to train the style transfer network, whereas our method does not require real examples of our object of interest. 
Our work leverages image synthesis techniques and physics-based rendering concepts but adapts them for shipwreck segmentation in side scan sonar imagery. We also contribute a novel method for simulating shipwreck debris using deformation fields, which we use as a proxy learning task for our method. 
\subsection{Domain Adaptation}
While simulation allows us to generate datasets with more examples of our target class, models trained on simulated or synthetic data will still suffer at test time due to the sim-to-real gap. Domain adaptation methods have been proposed to modify a network trained in a source domain (e.g. simulated data) to perform inference in a target domain (e.g. real data) \cite{adda2,cycada,hrda}. Relevant to this work is Unsupervised Domain Adaptation (UDA), where there are no object labels in the target domain~\cite{sepico, hrda}. Recently, Hoyer et. al proposed HRDA, a state-of-the-art UDA method for adaptation across different object sizes \cite{hrda}. Still, unsupervised domain adaptation requires a representative set of target objects to be present in both synthetic and real data for training, which may not be possible for rare targets. 
% \subsection{}
\subsection{Zero-shot Segmentation} 
Recently, zero-shot segmentation has been proposed to perform segmentation of novel objects at test time that were unseen during training \cite{zs3net, cagnet, kirillov2023segany}. Zero-shot segmentation generalizes segmentation learned on seen classes from a training dataset to unseen classes during test time. However, in our situation, we have access to labeled objects in a specific \textit{domain} and wish to generalize to another domain. The classes remain the same. A relatively new field called zero-shot unsupervised domain adaptation explores the problem of adapting a model learned in a specific domain to another domain without any examples from the target domain. Models like Prompt-driven Zero-shot Domain Adaptation (PODA) are prompted by text descriptions of the target domain and are able to adapt features learned in the source domain \cite{zuda}. Our method does not require prompts or text description, and instead leverages synthetic data for sim-to-real transfer. Thus, our method addresses a unique problem of zero-shot sim-to-real transfer for segmentation.
% Also related is Unsupervised Open Set Domain Adaptation (OSDA)~\cite{DBLP:journals/corr/abs-2110-04111, openset2}. Most OSDA algorithms deal with the case of novel classes that were not present in the source domain appearing in the target domain training data to learn to classify novel objects as an ``unknown" class. Note that this still requires the presence of objects of interest appearing in the target domain during training. 
%However, our case is the inverse, with two classes in the source domain training data $\{shipwreck, terrain\}$, but only one in the target domain training data $\{terrain\}$. 
%UDA methods work well to close the sim-to-real gap for certain problems. We investigate the performance of a zero-shot UDA method (PODA) and a traditional UDA method (HRDA) and compare them to our method. 
% We instead focus on developing inductive biases to improve shipwreck segmentation given access to only synthetic images of shipwrecks. 
 \subsection{Deep Anomaly Detection and Salient Object Detection} 
%\textbf{Deep Anomaly Detection: }
Salient object detection (SOD) aims to segment the primary object in an image from the scene. Although SOD networks like InSPyReNet do not explicitly consider sim-to-real gaps, they are designed to perform segmentation on novel objects and scenes \cite{kim2022revisiting}.
Prior works on anomaly detection also have a natural utility for this problem: they can generalize to the diverse set of anomalies encountered during test-time.
%Prior work on anomaly detection is also relevant to our work, as anomaly detection focuses on the problem of detecting objects or features at test time that are different from those seen during training. 
Unsupervised anomaly detection methods train only on a normal set of images and identify anomalies in test images by producing an anomaly score at the image or pixel level \cite{patchcore, dest, reiss_panda_2021}. DeSTSeg is an anomaly detection method that uses a student-teacher paradigm to segment anomalous regions of images \cite{dest}. DeSTSeg first generates synthetic anomalies then trains a supervised segmentation network. However, the synthetic anomalies are restricted to small cracks and defects often found in an industrial setting, not the complex shapes of shipwrecks as in our problem. The current state-of-the-art method, PatchCore, uses a memory bank to store normal features \cite{patchcore}. Finally, PatchCore provides an anomaly score by reporting a distance metric between test features and a subsampled memory bank. Our method, STARS, leverages anomalous features but refines them in a manner relevant to the segmentation task. We show that this ability to identify anomalous features during test-time is useful for minimizing the sim-to-real gap in a segmentation task.

%Anomaly detection has a natural utility for this problem: images of normal terrain are much easier to capture in large quantities than shipwreck anomalies. Although anomaly detection algorithms exhibit impressive performance in controlled environments like factory production lines, current methods suffer from high false positive rates when trained on unstructured data from field robotic surveys. 
%STARS uses anomalous features and refines them in a manner relevant to the segmentation task. We show that this ability to generalize to unseen features is useful for minimizing the sim-to-real gap. 

% We hypothesize that the ability for a network to identify anomalous areas is useful for the task of shipwreck detection given the diversity of debris fields. Our work incorporates anomaly detection in the form of a supplemental branch that is fused into the final segmentation output. 

% \begin{figure*}[t]
%     \centering
%     \includegraphics[width=7in]{synths}
%     \caption{(from left to right) Real side scan sonar imagery of the terrain, simulated terrain imagery, real shipwreck imagery, simulated shipwreck imagery. Note the randomization of blotches in simulated terrain imagery.}
%     \label{synthdata}
% \end{figure*}

\section{Technical Approach}
STARS has two main components: synthetic data generation (Fig. \ref{synth_overview}) and network development and training (Fig.~\ref{arch}). %First, we generate synthetic side scan sonar data for training using physics-based rendering. We introduce a simple but effective method for fracturing our ship using a deformation field and composite it into real terrain images. Next, we train our segmentation network completely in simulation. Finally, we perform inference on real side scan sonar images with no additional fine-tuning.  
  
% Training and inference are split into different parts. The inputs to the system are real terrain data and synthetic shipwreck data of resolution $(H,W)$. The final output of the anomaly inference is binary predictions for patch regions of size $(p_h, p_w)$. First, images are passed through an image encoder to produce a feature map $f \in \mathbb{R}^{n \times h \times w}$. Then, the feature map $f$ is used for both domain adaptation and compactness losses. During inference, normal image patches are passed through the trained image encoder to produce features. An anomaly detection model called an Isolation Forest is used to predict anomalies \cite{4781136}. 

\subsection{Synthetic Sonar Image Generation \label{sy}} 
%Although rendering side scan sonar images with ray-tracing is common \cite{sssrend}, to the best of our knowledge no one has incorporated deformation fields to produce diverse shipwreck sites. \\ \\
%Figure \ref{synth_overview} illustrates our pipeline for generating synthetic side scan sonar imagery. %First, 3D meshes of ships are collected from TurboSquid, imported into Blender, and ray-traced \cite{turbosquid}. Next, the output of ray-tracing is rendered according to the SONAR equations after parameter randomization \cite{Sanders1976AnIT}. Then, the ship is fractured in the image space using optical flow. Finally, the synthetic segmentation mask is used to composite the shipwreck onto real terrain. 
% \subsubsection{Simulated Environment Setup}
%\textbf{Simulated Environment Setup:} 
We generate synthetic side scan sonar data for training using physics-based rendering. We extended BLAINDER to develop a side scan sonar simulation system within the Blender graphics environment~\cite{s21062144, blender}, modeled after the side scan sonar used for real experiments.
% The Edgetech 2205 side scan sonar used in our field experiments has a 3dB beamwidth of $\theta_b =0.36^\circ$. Rays are traced accordingly from the sensor origin within a range of azimuth $\Theta \in [-\frac{\theta_b}{2}, \frac{\theta_b}{2}]$ and elevation angles $\Phi \in [90^\circ, 180^\circ]$. 
Our simulation environment consists of 3D meshes of ships downloaded from TurboSquid~\cite{turbosquid}. We randomized acoustic reflectance, ship location, ship scale, ship orientation.%\\ \\
% \subsubsection{Side Scan Sonar Rendering}
%\textbf{Side Scan Sonar Rendering: }
% To convert ray traced points in the sensor frame to a side scan sonar image, we use a method similar to that of \cite{sssrend}. The sonar has a range limit of $d_{max}$ meters and meters per pixel resolution of $\Delta =\frac{d_{max}}{W}$. A single simulated sonar ping is defined as the set of all the rays $\mathcal{R}$ traced through azimuth and elevation angle ranges, ($\Theta, \Phi$). A single ray from Blender $r_i \in \mathcal{R}$ provides material information, distance to intersection $d_i$, and normal information $n_i$. 
  \begin{figure*}[tp]
    \centering
    \includegraphics[width=\textwidth]{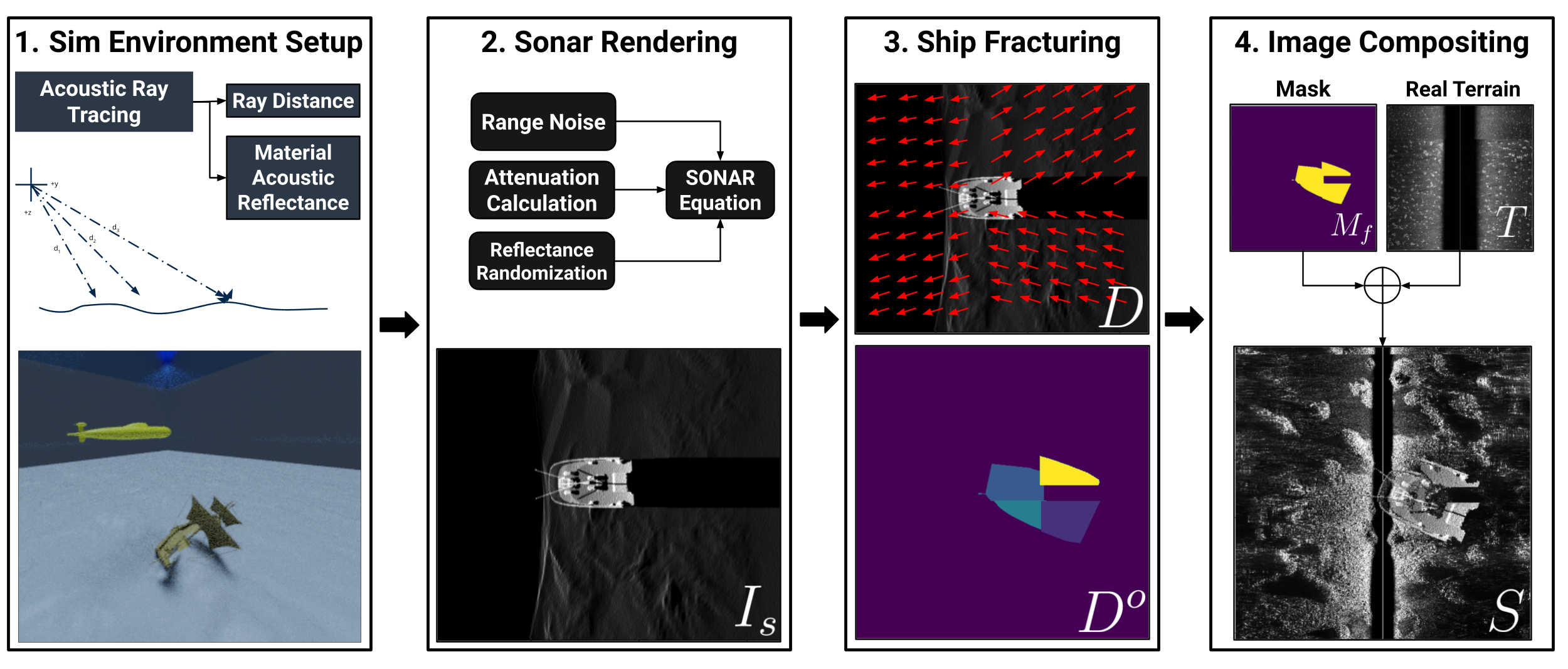}
    \caption{Shipwrecks are generated in a graphics environment then rendered into sonar images based on the SONAR equation. Then, they are fractured using deformation fields and composited onto real terrain images. }
    \vspace{-4mm}
    \label{synth_overview}
\end{figure*}

To calculate the intensity of the sonar return for a given ray $r_i$, we account for the propagation of sound in decibels underwater using the SONAR equation~\cite{Sanders1976AnIT}:
\begin{equation}
\label{sonar}
    RI_i = SL_i - 2TL_i + TS_i
\end{equation}
% \begin{figure}[t]
%     \centering
%     \includegraphics[width=3.5in]{angles.png}
%     \caption{ }
%     \label{fracture}
% \end{figure}

% Each horizontal pixel location $p$ of a side scan sonar image of resolution $(H,W)$ measures the intensity of all acoustic returns in a range $\mathcal{B}(p) = [d_p - \Delta/2, d_p + \Delta/2]$ meters from the sensor.
\noindent where $RI_i$ is the intensity of sound returned to the sensor, $SL_i$ is the source level of the emitted acoustic pulse, and $TL_i$ is the transmission loss from sound propagation through water such that $TL_i = 10\text{log}_{10}(d_i)$, where $d_i$ is the distance of the return. $TS$ is the target strength, or sonar cross section of the object imaged. We are able to produce both side scan sonar images and segmentation masks from our simulation environment.%\\ \\

We introduce a novel method for simulating diverse debris fields in side scan sonar images. This process is illustrated in Fig. \ref{synth_overview}. A deformation field, or optical flow field, dictates how pixels are translated in relation to the original image. The origin is at the center of the ship and the ship is split into four pieces. The field is randomly generated, but all pixels within a given quadrant experience the same field. The field $D \in \mathbb{R}^{H\times W \times 2}$ is parameterized by magnitude and direction $(r, \theta)$ for each pixel. The magnitude $r \in [0, r_{max}]$ is divided into $N_r = 10$ discrete values, whereas the direction $\theta \in [0, 2\pi]$ is divided into $N_\theta = 20$ discrete values. This is used to create a one-hot representation of the magnitude and direction $D^{mag} \in \mathbb{R}^{H \times W \times N_r}, D^{ang} \in \mathbb{R}^{H \times W \times N_\theta}$. These are concatenated to create $D^o = D^{mag} \oplus D^{ang} \in \mathbb{R}^{H \times W \times D_{def}}$, where $D_{def} = N_r + N_\theta = 30$. Then, the field is applied to split the ship image $I_s$ in an arbitrary pattern and produce the fractured image $I_f$. Let $I_s(u,v)$ represent the image value at pixel location $(u,v)$ and $(r, \theta) = D(u,v)$ represent the deformation field at the same location, then
\begin{equation}
I_f(u+rcos(\theta),v+rsin(\theta)) = I_s(u,v)
\end{equation}

\noindent The synthetic segmentation mask $M$ is also fractured with $D$ to produce fractured segmentation map $M_f$. The final synthetic image $S$ is created by compositing onto a real terrain image $T$. It is reasonable to assume access to unlabeled, real terrain images $T$ because they are publicly available and collected during routine surveys of bodies of water. 
% with an element-wise product $\odot$. 
% \begin{equation}
% S = [M_f \odot I_f] + [(1-M_f)\odot T]
% \end{equation}
% It is reasonable to assume access to real terrain images $T$ because they are publicly available and collected during routine surveys of bodies of water. This method of fracturing the ships is more convenient compared to fracturing in 3D because it avoids re-rendering the sonar images and we can expand upon our basic deformation field in the future. %need citation for this https://www.ngdc.noaa.gov/mgg/bathymetry/hydro.html 

% \begin{table*}[]
% \centering
% \caption{Randomization parameters for side scan sonar data generation \label{randparam}}
% \begin{tabular}{l|lllllllll}
% Parameter & Reflectance & Terrain Elevation & Ship X  & Ship Y & Ship Z  & Ship Scale & Ship Pitch    & Ship Roll     & Ship Yaw        \\ \hline
% Range     & 0.2-0.5              & ANT Randomization & {[}-10 10{]}  & {[}-10, 10{]} & {[}-10, 10{]} & {[}1, 5{]} & {[}-15, 15{]} & {[}-90, 90{]} & {[}-180, 180{]}
% \end{tabular}

% \end{table*}

\subsection{Network Architecture}
Our key insight with STARS lies in identifying the root causes of our sim-to-real gap and explicitly developing architectural inductive biases that mitigate them. First, we observe that any object underwater (especially shipwrecks) can experience extreme environmental degradation and destruction, resulting in fracturing and deformation. To address this, we introduce a deformation prediction proxy task supervised by our synthetic data that makes our network aware of fracturing and the spatial relationship between fragments and the whole ship. Secondly, we find that the appearance of real shipwrecks may differ from synthetic training data due to their high natural diversity. To address this challenge, our anomaly volume informs STARS of generally anomalous features in the scene that may aid the segmentation task. 

\begin{figure*}[t]
    \centering
    \includegraphics[width=\textwidth]{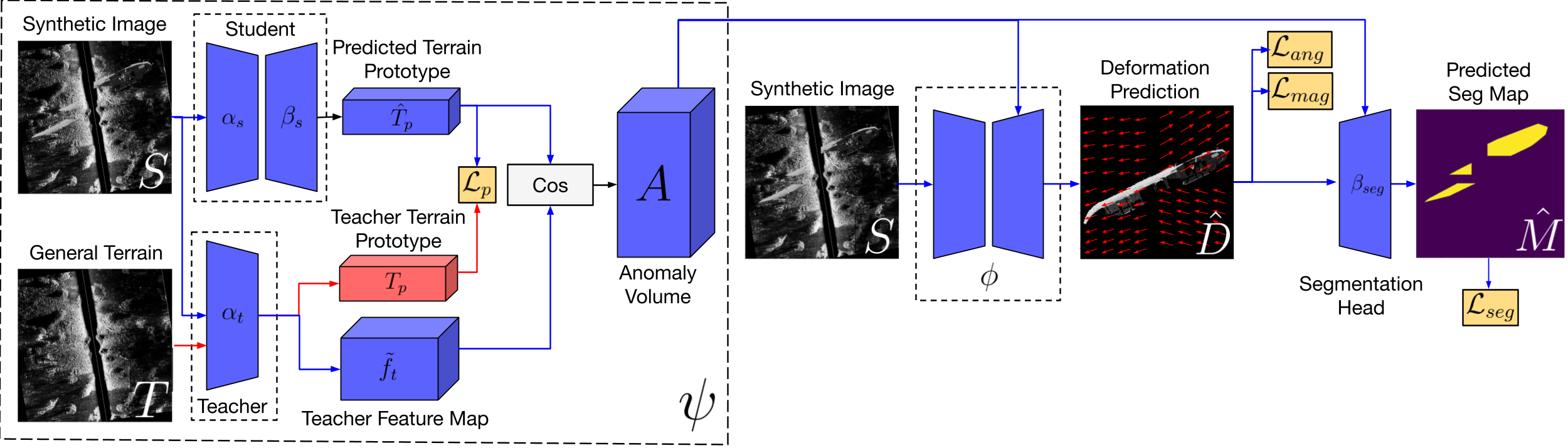}
    \caption{STARS network architecture. The anomaly prediction network $\psi$ produces anomaly volume $A$ by taking the cosine distance between the student's predicted terrain prototype $\hat{T}_p$ and the teacher's feature map of the synthetic image $\tilde{f}_t$. Then, the anomaly volume $A$ is used within the deformation prediction network $\phi$ to produce deformation field $\hat{D}$. The outputs of $\phi$ and $\psi$ are fused by $\beta_{seg}$ to produce a segmentation prediction. During inference, the teacher produces $\tilde{f}_t$ and the anomaly volume can be computed. All modules in {\color{blue} blue} are used for training and inference, while modules in {\color{red}{red}} are needed only for training. Best viewed in color and zoomed in.}
    \label{arch}
    \vspace{-4mm}
\end{figure*}

The detailed network architecture for STARS is shown in Fig.~\ref{arch}. Given an image $I \in \mathbb{R}^{H \times W \times 3}$, we wish to produce a segmentation map $\hat{M} \in \mathbb{R}^{H \times W \times 1}$ that segments two classes: $\{terrain, shipwreck\}$. We use a deformation prediction network $\phi:~\mathbb{R}^{H \times W \times 3} \rightarrow \mathbb{R}^{H \times W \times D_{def}}$ and anomaly prediction network $\psi:~\mathbb{R}^{H \times W \times 3} \rightarrow \mathbb{R}^{H \times W \times D_l}$. The outputs are fused by a lightweight decoder $\beta_{seg}$ to produce segmentation prediction $\hat{M}.$ Note that although deformation prediction, optical flow estimation \cite{Teed2020RAFTRA, dai2017deformable} and anomaly detection \cite{patchcore, reiss_panda_2021} have been independently explored in the vision community, they have not been combined in this manner to produce segmentation outputs in a zero-shot sim-to-real transfer setting for sonar data. Please refer to supplementary material for layer-level network architecture details. 

% The inductive biases our method develops are 1) the ability to recognize and discriminate deformations on ships and 2) the ability to identify and refine anomalous features concerning our target class of shipwreck.  

\subsubsection{Anomaly Prediction Network $\psi$}
Inspired by \cite{dest}, our method uses a student-teacher paradigm to produce an anomaly volume. \cite{dest} detects anomalies by computing the cosine distance between student and teacher features in the same spatial location. Our student network instead produces a single terrain prototype $\mathbb{R}^{1 \times 1 \times C}$ that summarizes the general terrain in an image.  By forcing the network to summarize the entire terrain with a single terrain prototype, we create a natural bottleneck for more efficient representation of terrain patterns. Then, our method computes the cosine distance between a terrain prototype and the teacher's entire feature map.

Our anomaly prediction network $\psi$ is composed of student encoder $\alpha_s$, student decoder $\beta_s$, and teacher encoder $\alpha_t$. The teacher network is frozen and does not receive gradient updates. First, a synthetic image $S$ is passed through the student encoder $\alpha_s$ and decoder $\beta_s$. The decoder produces feature maps $f_s(i), i \in [1,D_l]$ at varying resolutions. Then, a global average pooling (GAP) layer is used on the spatial dimensions of $f_s(i)$ to produce a student terrain prototype $\hat{T}_p(i)$. 
% \begin{equation} 
% \hat{T}_p(i)= GAP[f_s(i)] 
% \end{equation}
This terrain prototype is supervised by the teacher encoder $\alpha_t$. Note that the teacher is fed the regular terrain image $T$ \textit{without any objects}. This is easily taken from the synthetic image generation phase. $\alpha_t(T)$ produces feature maps $f_t(i), i \in [1,D_l]$. Similarly, global average pooling is used to construct a teacher terrain prototype, $T_p(i)$.
% \begin{equation} 
% T_p(i)= GAP[f_t(i)] 
% \end{equation}
Finally, an Mean Squared Error (MSE) loss is used to supervise the student: 
 \begin{equation}
 \mathcal{L}_p = \sum_{i=1}^{D_l} ||\hat{T}_p(i) - T_p(i)||_2^2
 \end{equation}

In order to compute the anomaly map $A(i)$ at a given depth $i$, the teacher encoder is fed the synthetic image $S$. This produces a teacher feature map $\tilde{f}_t(i)$. The cosine distance between each $\tilde{f}_t(i)$ and $\hat{T}_p(i)$ is used as an anomaly score: 
\begin{equation}
A(i) = 1 - \frac{\hat{T}_p(i) \cdot \tilde{f}_t(i)}{|\hat{T}_p(i)||\tilde{f}_t(i)|}
\end{equation}

Intuitively, the terrain prototype will have smaller cosine distance when compared to terrain features but higher distance when compared to anomalous debris features. Since the input to the student is $S$ and supervised by $T_p(i)$, it will learn to ignore any objects and focus on summarizing the terrain effectively. Note that $T_p$ is not needed for inference and is only used for supervised training of the student.

\subsubsection{Deformation Prediction Network $\psi$} 
Forcing STARS to predict the deformation field $\hat{D}$ that turned intact image $I_s$ into fractured image $I_f$ implicitly teaches STARS to identify parts of a broken ship, then learn the spatial relation between the different pieces. The decoder also concatenates the anomaly maps at varying resolutions $A(i), i \in [1,D_l]$ to the skip connections. This allows the anomaly prediction to aid the deformation prediction. %: $\hat{D} = \phi(S, A) $.

We pose deformation prediction as a classification task, with discretized magnitude and phase components. \noindent $\hat{D} \in \mathbb{R}^{H \times W \times D_{def}}$ is composed of a magnitude $\hat{D}^{mag} \in \mathbb{R}^{H \times W \times D_r}$ and angle  $\hat{D}^{ang} \in \mathbb{R}^{H \times W \times D_\theta}$ prediction for each pixel. These predictions are supervised with one-hot ground truth $D^o$ using a cross entropy losses $\mathcal{L}_{mag}$ and $\mathcal{L}_{ang}$.
% \vspace{-8mm}
% \begin{multicols}{2}
%   \begin{equation}
%      \mathcal{L}_{mag} = -\sum_{c=1}^{D_r}\sum_{h=1}^{H}\sum_{w=1}^{W}D^{mag}_{hwc}log\hat{D}^{mag}_{hwc}
%   \end{equation}\break
%   \begin{equation}
%     \mathcal{L}_{ang} = -\sum_{c=1}^{D_\theta}\sum_{h=1}^{H}\sum_{w=1}^{W}D^{ang}_{hwc}log\hat{D}^{ang}_{hwc}
%   \end{equation}
% \end{multicols}
% \begin{equation}
% \mathcal{L}_{mag} = -\sum_{c=1}^{D_r}\sum_{h=1}^{H}\sum_{w=1}^{W}D^{mag}_{hwc}log\hat{D}^{mag}_{hwc}
% \end{equation}
% \begin{equation}
% \mathcal{L}_{ang} = -\sum_{c=1}^{D_\theta}\sum_{h=1}^{H}\sum_{w=1}^{W}D^{ang}_{hwc}log\hat{D}^{ang}_{hwc}
% \end{equation}

%($\langle  \cdot  \rangle_{H,W}$) 
\subsubsection{Segmentation Decoder $\beta_{seg}$} 
The two branches are fused with a $1\times 1$ convolutional decoder $\beta_{seg}$. All layers $A(i)$ are bilinearly interpolated to a resolution of $(H,W)$ and concatenated along the channel dimension to produce feature volume $F \in \mathbb{R}^{H \times W \times D_l}$. 
% \begin{equation}
% F = \hat{D} \oplus \bigoplus_{i=1}^{D_l}   \langle A(i) \rangle_{H,W} 
% \end{equation}
Finally, the segmentation map is given by $\hat{M} = \beta_{seg}(F)$. The segmentation output is supervised by a binary cross entropy loss, $\mathcal{L}_{seg}$. 
% \begin{equation}
%     \mathcal{L}_{seg} = -\sum_{h=1}^{H}\sum_{w=1}^{W}(M_f)_{hw}log(\hat{M})_{hw}
% \end{equation}
The final loss becomes 
\begin{equation}
\mathcal{L} = \mathcal{L}_{mag} + \mathcal{L}_{ang} + \mathcal{L}_{p} + \mathcal{L}_{seg}
\end{equation}

%\begin{figure*}[t]
%    \centering
%    \includegraphics[width=7in]{sites.png}
%    \caption{Examples of shipwrecks collected from field work experiments. Note the extreme variation in terrain, shipwreck size/shape, and levels of degradation. The variety makes this a challenging perception problem as networks must generalize to multivarious environments. Sites from the hard split are denoted with a *.}
%    \label{sites}
%\end{figure*}
\section{Experiments \& Results} 
\subsection{Datasets} 
To collect a real dataset, field tests were carried out using an Iver-3 AUV in Thunder Bay National Marine Sanctuary (TBNMS) in Lake Huron, Michigan. TBNMS has almost 100 known shipwreck sites of varying size, type, and wreck condition, making this an ideal location for dataset collection to validate the proposed approach for the application of shipwreck segmentation \cite{tbnms}. %The shipwrecks we surveyed were: \textit{Barge No. 1, E.B. Allen, Flint, Grecian, Haltiner, Heart Failure, Lucy, Monrovia, Reef, Rend, W.P Thew, Viator, Wilson, Montana}. The \textit{Reef} site includes an artificial reef to serve as a distractor site. The survey sites are shown in Figure (\ref{sites}).   
%\begin{figure}[t]
%    \centering
    %\includegraphics[width=3.5in]{sites.png}
%    \caption{Annotated Map of Thunder Bay National Marine Sanctuary with proposed survey regions. Red indicates known shipwreck sites in the area, green are known sites that we prioritized on our surveys, yellow are suspected sites. \label{sites}}
%\end{figure}
% All experiments were conducted with the Michigan Technological University Great Lakes Research Center Iver3 AUV equipped with an EdgeTech 2205 side scan sonar. 
% The maximum ranges used for our surveys were between 60-150 meters for each transducer.
% We executed both lawnmower patterns and Object Identification patterns (OID). This allowed us to collect imagery of large swaths of general terrain, as well as multiple views from different orientations of the same shipwreck with OID patterns.
Surveys were conducted over a two week period and produced images from 13 distinct shipwreck sites and 1 artificial reef that serves as a potential distractor object. The field work surveys resulted in 220 scans of terrain and shipwrecks. The raw scans are of variable height, but maintain a horizontal resolution of 1728 pixels. We take a sliding window of size $(1728, 1728)$ and generate images with vertical stride of 100 pixels. The resulting dataset has 861 images, including 312 images of terrain and 549 images of shipwrecks. We do not use the shipwreck images for training, and instead withhold them as a test set for evaluation only. % Characteristic sonar images of each site are shown in Fig. \ref{sites}.  
%The raw sonar data is timestamped with GPS coordinate estimates from dead-reckoning state estimation. Other relevant sensors include barometer, IMU, compass, and a forward looking sonar.  
%\subsection{Dataset Details}
%\subsection{Real Dataset}
%primary concern a reviewer could bring up: dataset size 
%\textbf{Dataset Overview:} 

Labeling side scan sonar images can be challenging because of shadowing effects, view-dependent acoustic artifacts, and complex debris fields. %Furthermore, it can be difficult for a labeler to build intuition about the 3D structure of objects through the sonar projection model. 
To ensure accurate ground truth, each shipwreck site was labeled with the help of an expert marine archaeologist from TBNMS %from the Michigan State Department of Natural Resources 
who has extensively studied and visually confirmed these wrecks by scuba diving.

We wish to emphasize that collecting side scan sonar imagery is extremely expensive and time consuming, even with autonomous systems. We recognize that our dataset, composed of 220 distinct scans, is orders of magnitude smaller than large scale benchmark datasets collected for optical deep learning. However, given the diversity of our dataset, we believe it is representative of sites seen during real deployment. Our TBNMS dataset provides useful insight into how well networks can generalize to real, challenging, and unstructured sonar surveys. Please refer to supplementary material for sample data from our field surveys. %Examples of the surveyed sites are shown in Figure (\ref{fig:sites}).%\\ 
%\textbf{Image Information:} 

%The raw scans are of variable height, but maintain a horizontal resolution of 1728 pixels. We take a sliding window of size $(1728, 1728)$ and generate images with vertical stride of 100 pixels. The resulting dataset has 861 images, including 312 images of terrain and 549 images of shipwrecks. We do not use the shipwreck images for training, and instead withhold them as a test set for evaluation only.  %\\
%\textbf{Label Information:} 

%Labeling side scan sonar images can be challenging because of shadowing effects, view-dependent acoustic artifacts, and complex debris fields. %Furthermore, it can be difficult for a labeler to build intuition about the 3D structure of objects through the sonar projection model. 
%To ensure accurate ground truth, each shipwreck site was labeled with the help of an expert marine archaeologist %from the Michigan State Department of Natural Resources 
%who has extensively studied and visually confirmed these wrecks by scuba diving. %marine archaeologists like the one we worked with are able to better interpret the sonar imagery and provide detailed labels for complex debris fields. 

% - side scan sonar imagery is hard to label normally 
% - had an expert label it and he has visually confirmed all these sites 

%\subsection{Synthetic Dataset}
For simulated data, we create a dataset of 10,000 images of resolution $(1728, 1728)$ using our generation process outlined in Section \ref{sy}. An 80/20 split was used to create training and validation sets respectively.
%Reimplemented baselines (Burguera and Bonin-Font \cite{jmse8080557} and Yang et. al \cite{baseline2}) were trained under the same conditions. 
% \begin{table}[t]
% \caption{Baselines and their access to various training data \label{access}}
% \tabcolsep=0.11cm
% \centering
% \begin{tabular}{l|ccc}
% \hline
% Method      & Real Terrain & Synth. Ships & Real Ships \\ \hline
% PatchCore \cite{patchcore}   & \cmark        & \xmark                    &  \xmark               \\
% HRDA \cite{hrda}       & \cmark            & \cmark                  & \xmark                \\
% PODA \cite{zuda}       & \cmark            & \cmark                  & \xmark                \\
% % Sim2Real    & \xmark              & \cmark                   &  \xmark               \\
% HRNetV2 \cite{hrnet} & \cmark            & \cmark                   &  \xmark               \\
% Yang et. al \cite{baseline2} & \cmark            & \cmark                   &  \xmark              \\
% Burguera \cite{jmse8080557} & \cmark            & \cmark                   &  \xmark              \\ \hline
% STARS (ours)  & \cmark            & \cmark                   &  \xmark         \\ \hline     
% \end{tabular}
% \vspace{-4mm}
% \end{table}
\vspace{-3mm}
\subsection{Baselines}

% Please add the following required packages to your document preamble:
% \usepackage[table,xcdraw]{xcolor}
% If you use beamer only pass "xcolor=table" option, i.e. \documentclass[xcolor=table]{beamer}

% \begin{table*}[t]
% \caption{Segmentation performance of our method compared to baselines. Metrics are averaged across 14 sites. \label{overall}}
% \centering
% \begin{tabular}{c|cccccc}
% Method   & PatchCore \cite{patchcore} & HRDA \cite{hrda} & HRNet \cite{hrnet} & Yang et. al \cite{baseline2} & Burguera and Bonin-Font \cite{jmse8080557}   & Our Method \\ \hline
% Average IOU     & 0.28      & 0.19 &   0.33          && &\textbf{0.49}       \\
% F1 Score & 0.43      & 0.29 &    0.47         & &&\textbf{0.63}      
% \end{tabular}
% \end{table*}

% \begin{figure*}
% \caption{Per-site IOU$_{ship}$ segmentation performance trained on \textbf{only} simulated data. Last column is averaged across all 14 sites.  \label{persite}}

%     \includesvg[width=\linewidth]{figures/per_site.svg}
% \end{figure*}

We chose a variety of baselines that perform segmentation while respecting the restriction on real training data. We evaluate STARS against PatchCore unsupervised anomaly detection \cite{patchcore}, HRDA unsupervised domain adaptation \cite{hrda}, PODA zero-shot domain adaptation, \cite{zuda}, HRNetv2+OCR \cite{hrnet, ocr} as a baseline method for no sim-to-real transfer, Yang et. al as a side scan sonar segmentation method \cite{baseline2}, and InSPyReNet \cite{kim2022revisiting} as a salient object detection baseline. Please refer to supplementary information for more details of implementation and baselines.
% The data access for baselines is summarized in Table (\ref{access}):

\begin{figure*}[t]
    \begin{center}
    \includegraphics[width=5in]{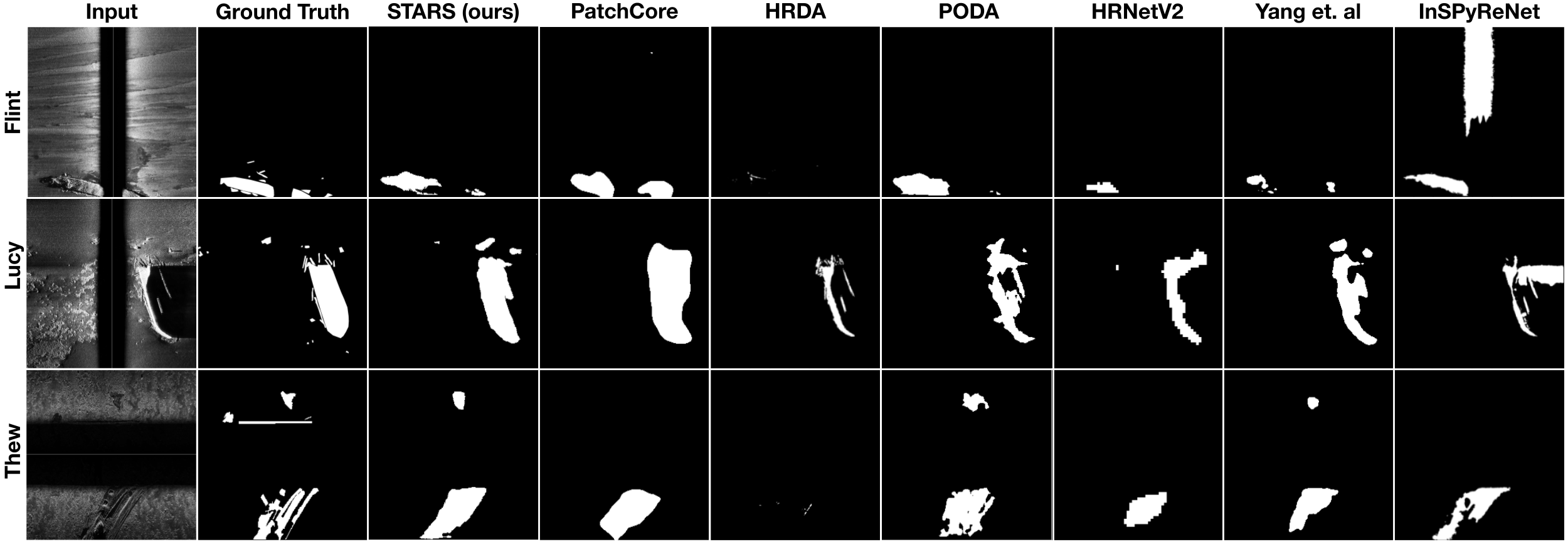}
    \end{center}
    \vspace{-3mm}
    \caption{Selected qualitative results from our method compared to baselines.
    % We insert red boxes to highlight areas where STARS performs better (they are not produced by the model). 
    It should be noted that some methods have a tendency to inaccurately over-segment or fail to segment debris from shipwrecks, resulting in lower performance. However, STARS consistently produces more accurate segmentation outputs. %In challenging sites like Thew and Viator, STARS is also able to segment smaller, more subtle debris where other methods fail completely. 
    Best viewed zoomed in. \label{qual}}
    \vspace{-4mm}
\end{figure*}

%\subsection{Evaluation Metrics for Segmentation}
%There are two classes of interest for our semantic segmentation task: $\{terrain, shipwreck\}$. We use IOU and F1 score to quantify the performance of the methods tested. We report the IOU of the shipwreck and terrain class, but terrain is considered a background class. Given the class imbalance between terrain pixels and shipwreck pixels (10:1 ratio respectively), we use F1 score instead of per-pixel accuracy. 
% \begin{equation}
% IOU = \frac{|A \cap B|}{|A \cup B|}, ~F1 = \frac{TP}{TP + \frac{1}{2}(FP + FN)}
% \end{equation}
% where $A$ is the network's prediction, $B$ is the ground truth pixels, $TP$ are true positives, $FP$ are false positives, and $FN$ are false negatives.

\subsection{Segmentation Performance} 
% \begin{table}[t]
% \centering 
%  \caption{Summarized segmentation performance of our method compared to baselines. Metrics are averaged across 14 sites. \label{overall}}
% \scalebox{0.7}{\begin{tabular}{l|cccc}
% \hline
% Method  & IOU$_{ship}$ $\uparrow$  & IOU$_{terr}$ $\uparrow$  & mIOU $\uparrow$ &  F1 Score $\uparrow$ \\ \hline
% PatchCore \cite{patchcore}                 & 0.28   & 0.91     &   0.60 & 0.43                           \\
% HRDA \cite{hrda}                           & 0.19   &  0.97       &  0.58 & 0.29                           \\
% PODA \cite{zuda} & 0.28 & 0.97 & 0.63 & 0.41 \\ 
% HRNetV2 \cite{hrnet}                         &0.35 &  0.97     &   0.66 & 0.48                          \\
% Yang et. al \cite{baseline2}              &   0.31  & \textbf{0.98}  &    0.65    & 0.48                               \\
% % Burguera \cite{jmse8080557} &         0.25      &   0.97       & 0.61      &       0.38                         \\ 
% InSPyReNet \cite{kim2022revisiting} & 0.33 &   0.97     & 0.65  &   0.45 \\ \hline  
% STARS (ours)  & \textbf{0.42} & \textbf{0.98} & \textbf{0.70} &\textbf{0.55} \\ \hline
% \end{tabular}}
% \vspace{-4mm}
% \end{table}

%\textbf{Quantitative Results. } 
Qualitative results from selected sites are shown in Figure (\ref{qual}). Note that none of the evaluated networks have seen a real shipwreck site during training, making segmentation at novel sites a very challenging task. All experiments use images of resolution (1728, 1728). 
% \newlength{\oldintextsep}
% \setlength{\oldintextsep}{\intextsep}
\setlength\intextsep{0pt}
\begin{wraptable}{r}{0.6\linewidth}
% \vspace{-4.9mm}
\caption{Summarized segmentation performance of our method compared to baselines averaged across 14 sites. \label{overall}}
\centering
\tabcolsep=0.11cm
\scalebox{0.7}{\begin{tabular}{l|cccc}
\hline
Method  & IOU$_{ship}$ $\uparrow$  & IOU$_{terr}$ $\uparrow$  & mIOU $\uparrow$ &  F1 Score $\uparrow$ \\ \hline
PatchCore \cite{patchcore}                 & 0.28   & 0.91     &   0.60 & 0.43                           \\
HRDA \cite{hrda}                           & 0.19   &  0.97       &  0.58 & 0.29                           \\
PODA \cite{zuda} & 0.28 & 0.97 & 0.63 & 0.41 \\ 
HRNetV2 \cite{hrnet}                         &0.35 &  0.97     &   0.66 & 0.48                          \\
Yang et. al \cite{baseline2}              &   0.31  & \textbf{0.98}  &    0.65    & 0.48                               \\
Burguera \cite{jmse8080557} &         0.25      &   0.97       & 0.61      &       0.38                         \\ 
InSPyReNet \cite{kim2022revisiting} & 0.33 &   0.97     & 0.65  &   0.45 \\ \hline  
STARS (ours)  & \textbf{0.42} & \textbf{0.98} & \textbf{0.70} &\textbf{0.55} \\ \hline
\end{tabular}}
\end{wraptable}
For quantitative results, we note that there are two classes of interest for our semantic segmentation task: $\{terrain, shipwreck\}$. We use Intersection over Union (IOU) and F1 score to quantify the performance of the methods tested. We report the IOU of the shipwreck and terrain class, but terrain is considered a background class. Given the class imbalance between terrain pixels and shipwreck pixels (10:1 ratio respectively), we use F1 score instead of per-pixel accuracy. The overall segmentation performance of our method is shown in Table (\ref{overall}). The IOU for both classes and F1 Score were calculated for each site then averaged across all sites. Our method consistently outperforms baselines, achieving \textbf{ 0.42 IOU$_{ship}$}. Additional results including detailed per-site segmentation performance are shown in supplementary material. 

%Our method consistently outperforms baselines, achieving a superior\textbf{ 0.42 IOU$_{ship}$}. On certain sites like \textit{Haltiner, Heart Failure, and Monrovia}, other methods like HRDA and HRNet perform better. However, on most sites like \textit{Barge \#1, Montana, Thew}, our method is able to produce more effective segmentation maps. Performance on \textit{Reef} is poor overall as the reef is a challenging distractor that looks similar to shipwrecks. %\\
%\textbf{Qualitative Results.} Qualitative results from selected sites are shown in Figure (\ref{qual}). Note that none of the evaluated networks have seen a real shipwreck site during training, making segmentation at novel sites a very challenging task. 

\subsection{Ablation Studies \label{abs}} 

\begin{wraptable}{r}{0.6\linewidth}
% \vspace{-4.95mm}
\centering
\caption{Network and Synthetic Data Ablation Studies \label{nab}}
\scalebox{0.7}{\begin{tabular}{c|cc|cc|cc}
\hline
Model & DB & AB &RT&SF & IOU$_{ship} \uparrow$ & F1 Score $\uparrow$ \\ \hline
No DB, No AB           &   \xmark                 &    \xmark & \cmark     & \cmark         &  0.26 & 0.38  \\
AB Only         &   \xmark    &  \cmark  & \cmark     & \cmark                       &  0.39 & 0.45   \\
DB Only         &   \cmark    &  \xmark   & \cmark     & \cmark                        & 0.36 & 0.45     \\ \hline
No RT, No SF           & \cmark     & \cmark &   \xmark                 &    \xmark                  &  0.24 &  0.33   \\
SF Only           & \cmark     & \cmark &   \xmark    &  \cmark                        &  0.12 &  0.19   \\
RT Only           &  \cmark     & \cmark &  \cmark                 &    \xmark                  & 0.36 & 0.49     \\ \hline
STARS (ours)   &  \cmark     & \cmark       &   \cmark           &   \cmark            & \textbf{0.42} & \textbf{0.55}    \\   \hline
\end{tabular}}
\end{wraptable}
Our method relies on synthetic data generation and a specialized network architecture to exploit the generated data. To investigate the importance of each factor, we ablate the deformation branch (DB), anomaly branch (AB), composition using real terrain (RT), and ship fracturing (SF) in Table (\ref{nab}). After adding the deformation branch (DB) and anomaly branch (AB) in isolation, performance increases by 0.10 and 0.13 IOU$_{ship}$, respectively. With respect to synthetic data generation, we found that real terrain (RT) is essential to improved performance, and ship fracturing (SF) without real terrain decreases performance. %Finally, enabling all these contributions performs the best at 0.42 IOU$_{ship}$. 

\section{Conclusion \& Future Work}
We propose a synthetic data generation framework and novel network architecture termed STARS for zero-shot sim-to-real segmentation of shipwrecks in side scan sonar imagery. Our model shows a significant \textbf{20\%} improvement in $IOU_{ship}$ from state-of-the-art baselines with access to the same data. %, including unsupervised domain adaptation networks. %Ablation studies show the utility of our anomaly prediction network, which increases segmentation performance by 50\% IOU$_{ship}$ compared to a baseline model without this feature. In conjunction with our deformation prediction network, segmentation performance increases by 62\% IOU$_{ship}$ with these features. We note that although our method is applied to shipwreck detection, many objects underwater experience the same environmental degradation and fracturing process that our network is designed to handle. This motivates future work in expanding to multi-class segmentation involving other underwater objects. 
%We found that despite having access to only 10 real images of terrain, our model achieved 0.35 mIOU, performing as well as or better than baselines with access to all 312 diverse images. This means during deployment, only a small representative dataset of terrain images may need to be collected to train the network completely in simulation. 
 %This shows that in challenging low-data regimes like automated target recognition underwater, using sim-to-real methods like ours may be more performant and cost effective than collecting and training on real data. 
Both UDA and zero-shot UDA methods performed reasonably well, but struggled to adapt features learned from synthetic data to the target domain. Our work has the potential to significantly reduce the cost and effort needed to train deep learning models for sonar segmentation by removing reliance on real, labeled data. 

Future work will focus on improving performance on hard sites, including challenging distractor sites, such as the artificial reef site. We note that although our method is applied to shipwreck detection, many objects underwater experience the same environmental degradation and fracturing process that our network is designed to handle. This motivates future work in expanding to multi-class segmentation involving other underwater objects. %A key aspect of achieving this is increasing the ability of the network to detect objects that blend in with the environment, relying more on shape cues than misleading acoustic reflectance properties. 
 %Improving performance on sites like \textit{Reef} may be achieved by including distractors in the data. 
 
%Our work has potential to significantly reduce the cost and effort needed to train deep learning models for object detection in side scan sonar by removing the reliance on real, labeled data. This will enable more resilient autonomous underwater systems that can operate in unstructured and unseen environments.   
\section{Acknowledgements}
We would like to thank Professor Timothy Havens, Professor Guy Meadows, Jamey Anderson and Chris Pinnow of the Great Lakes Research Center at Michigan Technological University for IVER-3 AUV data collection, and Thunder Bay National Marine Sanctuary for supporting field experiments. This work is supported by a University of Michigan Robotics Department Fellowship and by the NOAA Ocean Exploration program under Award \#NA21OAR0110196.

\bibliography{egbib}
\end{document}